\documentclass[runningheads]{llncs}
\usepackage[T1]{fontenc}
\usepackage{graphicx}
\usepackage{bbding}
\usepackage{booktabs} 
\usepackage{array}
\usepackage{multirow}
\usepackage{amsmath}
\usepackage[hidelinks]{hyperref}
\usepackage{color}

\usepackage{orcidlink}
\renewcommand{\orcidID}[1]{\,\orcidlink{#1}}

\begin{document}
\title{ACAT: A Collaborative Platform for Efficient Aspect-Based Sentiment Dataset Annotation}
\titlerunning{ACAT: ABSA Collaborative Annotation Tool}
\author{Ana-Maria Luisa Mocanu$^1$\orcidID{0009-0005-2646-4086} \Envelope \and
Ciprian-Octavian Truică$^{1,2}$\orcidID{0000-0001-7292-4462}
 \and 
Elena-Simona Apostol$^1$\orcidID{0000-0001-6397-4951}}

\authorrunning{A. Mocanu et al.}

\institute{$^1$National University of Science and Technology POLITEHNICA Bucharest, Romania\\
$^2$Academy of Romanian Scientists, Ilfov 3, Bucharest, 050044, Romania
\email{\{ana\_maria.mogoase, ciprian.truica, elena.apostol\}@upb.ro}\\
}

\maketitle

\begin{abstract}
Aspect-Based Sentiment Analysis (ABSA) requires high-quality datasets to train reliable models. However, existing annotation tools treat output as flat files, leaving researchers to manually consolidate multi-annotator data, reconstruct relational structures, and compute reliability metrics through custom scripts. This paper introduces ACAT (\textbf{A}spect-based sentiment analysis \textbf{C}ollaborative \textbf{A}nnotation \textbf{T}ool), a web-based platform natively supporting four ABSA workflows: (1)~Aspect-Category Sentiment Analysis, (2)~Clause-Level Segmentation, (3)~Aspect-Term Sentiment Analysis with character-level position tracking, and (4)~Aspect Sentiment Triplet Extraction with dual span offset preservation. Its core contribution is an automated Extract, Transform, Load (ETL) pipeline that aligns collaborative annotations and computes Inter-Annotator Agreement (IAA) metrics directly at export, yielding training-ready datasets. In a preliminary validation on $1{,}002$ restaurant reviews with two annotators of differing expertise, ACAT achieves a median annotation time of $31.58$~seconds and a raw IAA ranging from $0.78$ to $0.86$ across all tasks.

\end{abstract}

\section{Introduction} \label{sec:intro}
Sentiment Analysis has evolved from document-level polarity classification to Aspect-Based Sentiment Analysis (ABSA)~\cite{pontiki-etal-2014-semeval,Apostol2025}, encompassing multiple advanced tasks. Although Large Language Models (LLMs) have improved, human-annotated \emph{gold standard} datasets remain essential for reliable evaluation~\cite{aldeen2023chatgpt}. Yet the data management ecosystem for ABSA datasets remains fragmented: generic tools treat annotation as flat files, leaving data integration, multi-annotator merging, and IAA computation as the researcher's burden.

This paper introduces \textbf{ACAT}, addressing the following research questions: 
\textbf{(RQ1)}~How can the platform support ABSA knowledge extraction without complex pre-configurations? 
\textbf{(RQ2)}~What architectural solutions eliminate the data consolidation bottleneck in collaborative environments? 
\textbf{(RQ3)}~How can the annotation schema capture implicit or latent semantics usually lost in standard tools? 
\textbf{(RQ4)}~How can IAA be validated automatically without external post-processing scripts? 
\textbf{(RQ5)}~Why are direct one-to-one metric comparisons between existing solutions unreliable?

ACAT is deployed as a Docker container (PostgreSQL, Python Flask, Vanilla JavaScript). We validate it on $1{,}002$ restaurant reviews, achieving a median annotation time of 31.58~seconds and raw IAA up to 0.86. The contributions are:\\
\noindent\textbf{(C1) Unified Multi-Task Knowledge Extraction.} Native support for four ABSA architectures without pre-annotation configuration: ACSA~\cite{pontiki-etal-2014-semeval}, Clause-Level~\cite{clause-na-2015}, ATSA~\cite{pontiki-etal-2014-semeval}, and ASTE~\cite{peng2020knowing}.\\
\noindent\textbf{(C2) Automated ETL and Data Consolidation.} Row-level alignment of multi-user data with natively computed IAA metrics (Cohen's Kappa~\cite{cohen1960coefficient}, Fleiss' Kappa~\cite{fleiss1971measuring}, Macro F1~\cite{Truica2017}).\\
\noindent\textbf{(C3) Implicit Semantic Modeling.} An Implicit Toggle captures latent semantics when explicit target terms are absent~\cite{sun2025text}.\\
\noindent\textbf{(C4) Data Governance.} A high-speed double-click interaction model with built-in time tracking and hierarchical supervision.

\section{Related Work}\label{sec:related}
While ABSA algorithms advance rapidly, annotation infrastructure remains underdeveloped~\cite{colucci2024text}. General-purpose tools such as Doccano~\cite{doccano}, LightTag~\cite{perry2021lighttag}, and Label Studio~\cite{labelstudio} handle flat sequence labeling but export nested JSONs requiring custom scripts for ABSA task reconstruction and multi-annotator consolidation. Heavyweight environments like BRAT~\cite{stenetorp2012brat} support relational tasks and offer built-in curation modules, but impose click-heavy graph-drawing that introduce cognitive load. Offline tools such as YEDDA~\cite{yang-etal-2018-yedda} and ASQE-DPT~\cite{hua2025edurabsa} optimise local speed but lack centralized collaboration architectures, leading many ABSA datasets to rely on bespoke single-use pipelines~\cite{saeidi2016sentihood}. Finally, capturing implicit aspects remains challenging~\cite{li2021implicit,sun2025text}: while neural models encode implicit meaning~\cite{li2021implicit}, traditional tools restrict annotations to visible text spans. ACAT addresses all these gaps with native ABSA support, automated ETL, an implicit toggle, and a streamlined double-click interaction.

\section{Annotation Structure and Export}\label{sec:annotation}

\begin{figure}[t]
\centering
\includegraphics[width=0.92\columnwidth]{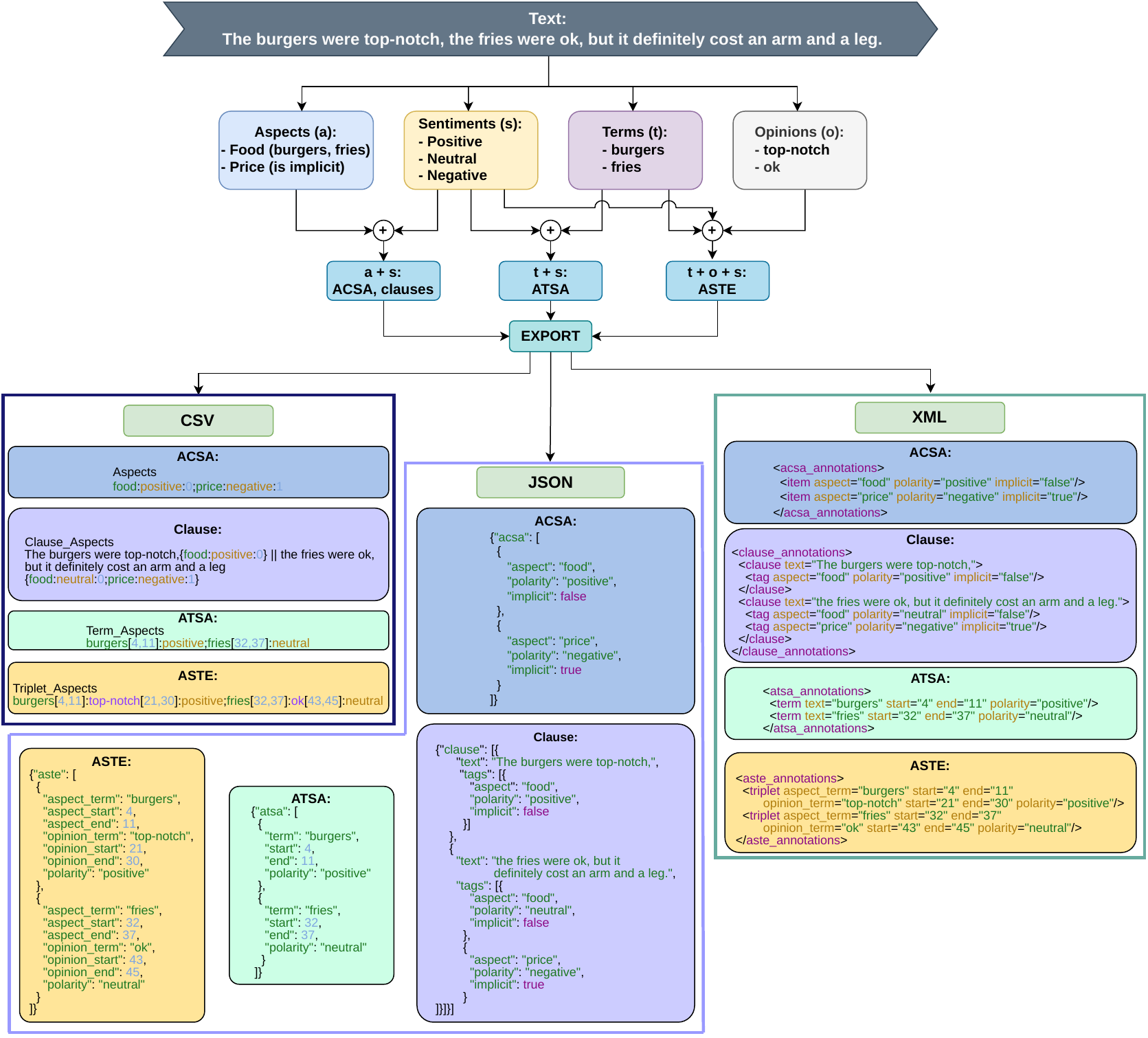}
\caption{ACAT export architecture. Linguistic variables $(a, s, t, o)$ are composed into four tasks of increasing granularity (top), with their CSV, JSON, and XML serialisations shown below, including character-level offsets and the implicit flag.}
\label{fig:export_system}
\end{figure}

ACAT supports four annotation granularity levels, each serialised as coordinate-based strings for storage and transformed into CSV, JSON, and XML at export. We employ a discrete 3-class polarity system $P = \{\textsc{Positive}, \textsc{Negative}, \allowbreak \textsc{Neutral}\}$~\cite{pang2009opinion,Petrescu2025}. Figure~\ref{fig:export_system} illustrates all tasks using our running example.
\footnote{The export example is available at \url{https://ronlp-clarin.upb.ro/acat.html}.}

The export logic composes four linguistic variables, Aspect Category~($a$), Sentiment Polarity~($s$), Aspect Term~($t$), and Opinion Word~($o$), into task-specific outputs: \textbf{ACSA \& Clause-Level ($a{+}s$)} maps latent categories to polarity labels; \textbf{ATSA ($t{+}s$)} combines explicit terms with character-level offsets; \textbf{ASTE ($t{+}o{+}s$)} links aspect and opinion spans, preserving dual offset pairs.

\textbf{ACSA} serialises annotations as \texttt{category:polarity:implicit}. The \textbf{Implicit Toggle}~(C3) extends the category-polarity pair to a formal triplet $(c, p, i)$, $i \in \{0,1\}$, flagging cases where a category is discussed without an explicit surface term~\cite{pontiki-etal-2014-semeval}. ACAT supports an open-world paradigm: annotators can dynamically extend the aspect taxonomy during active sessions.

\textbf{Clause-Level Sentiment Analysis} segments complex sentences into Elementary Discourse Units (EDUs), minimal, clause-level text spans each expressing a single sentiment, via an interactive splitter, resolving the mixed-sentiment problem. Clauses are delimited by \texttt{||} in CSV; JSON uses a \texttt{clause} array pairing each segment with a \texttt{tags} array.

\textbf{ATSA} captures character-level offsets from the annotator's text selection, e.g.\ \texttt{burgers[4,11]:positive;fries[32,37]:neutral}~\cite{Apostol2025}.

\textbf{ASTE} extracts the complete relational triplet $(t, o, s)$~\cite{peng2020knowing}, serialising dual offset pairs, e.g.\ \texttt{burgers[4,11]:top-notch[21,30]:positive;\allowbreak fries[32,37]
:ok[43,45]:neutral}. Duplicate entries are automatically detected and rejected.

\textbf{Collaborative ETL Pipeline.}
During export, ACAT performs automatic row-level alignment, dynamically joining dataset records with user-specific annotation layers and resolving sparse matrices where annotators were assigned disjoint subsets. Let $R = \{r_1,\ldots,r_n\}$ be the review instances and $U = \{u_1,\ldots,u_m\}$ the annotators. The pipeline generates a unified feature matrix $M$ where $M_{i,j}$ contains the annotation by $u_j$ for $r_i$. Three IAA metrics are computed via scikit-learn and statsmodels and embedded in all export formats (answering RQ4).
Cohen's Kappa~\cite{cohen1960coefficient} 
is applied for two-annotator datasets; Fleiss' Kappa~\cite{fleiss1971measuring}
is automatically selected for three or more annotators; Macro F1~\cite{Truica2017} 
accounts for class imbalance across all sub-tasks.

\section{Analysis and Discussions} \label{sec:results}
Two annotators each processed $N{=}100$ reviews per task, sampled from a corpus of $1{,}002$ restaurant reviews. ACAT's dataset-sharing ensured identical review sets per task. One annotator had prior ABSA experience (A1), while the other was a novice (A2), a deliberate choice to assess platform intuitiveness across expertise levels.

Table~\ref{tab:time_analysis} summarises temporal and agreement results. The overall median annotation time is 31.58~seconds (mean 37.77~s). A1 completed 58-67\% of reviews within 30~seconds; A2 completed 32-37\%. Overall, 49\% of annotations were completed within 30~seconds, suggesting that the double-click interaction model reduces friction for typical inputs. The long tail (maximum 164~s, standard deviation up to 30.35~s) reflects the complexity of dense reviews requiring open-world aspect extension.

\begin{table}[!htbp]
\centering
\caption{Temporal performance (seconds) and inter-annotator agreement metrics. \textbf{Agr.} is Raw Agreement (\% of identical labels between annotators). \textbf{Eff.} is efficiency (\% of reviews annotated in $\le 30$ seconds).}

\label{tab:time_analysis}
\setlength{\tabcolsep}{4pt}
\begin{tabular}{@{}ll rrrrr c rrr@{}}
\toprule
\multirow{2}{*}{\begin{tabular}[c]{@{}l@{}}\textbf{Task}\end{tabular}} & \multirow{2}{*}{\begin{tabular}[c]{@{}l@{}}\textbf{Ann.}\end{tabular}}  & \multicolumn{5}{c}{\textbf{Temporal Metrics (s)}} & & \multicolumn{3}{c}{\textbf{Agreement}} \\
\cmidrule{3-7} \cmidrule{9-11}
 &  & \textbf{Mean} & \textbf{Med.} & \textbf{Std.} & \textbf{Min} & \textbf{Max} & \textbf{Eff.} & \textbf{$\kappa$} & \textbf{F1} & \textbf{Agr.} \\
\midrule

\multirow{2}{*}{\begin{tabular}[c]{@{}l@{}}\textbf{ACSA}\end{tabular}} & A1 & 28.58 & 22.00 & 22.43 & 4 & 141 & 67\% & 0.60 & 0.59 & 0.86 \\
               & A2 & 41.81 &  36.50 & 27.43 & 9 & 164 & 37\% & ---  & ---  & ---  \\
\multirow{2}{*}{\begin{tabular}[c]{@{}l@{}}\textbf{Clause}\end{tabular}}& A1 & 35.05 & 29.00 & 23.70 & 5 & 128 & 58\% & 0.65 & 0.64 & 0.82 \\
                & A2 & 51.27 & 48.11 & 28.98 & 11 & 149 & 32\% & ---  & ---  & ---  \\
\multirow{2}{*}{\begin{tabular}[c]{@{}l@{}}\textbf{ATSA}\end{tabular}} & A1 & 27.98 & 21.50 & 22.58 & 3 & 107 & 66\% & 0.58 & 0.57 & 0.84 \\
                & A2 & 40.93 & 35.67 & 27.61 & 7 & 124 & 36\% & ---  & ---  & ---  \\
\multirow{2}{*}{\begin{tabular}[c]{@{}l@{}}\textbf{ASTE}\end{tabular}} & A1 & 31.07 & 22.50 & 24.82 & 5 & 127 & 62\% & 0.52 & 0.50 & 0.78 \\
                & A2 & 45.45 & 37.33 & 30.35 & 11 & 148 & 34\% & ---  & ---  & ---  \\
\bottomrule
\end{tabular}
 \vspace{-1.2em}
\end{table}

Raw agreement ranges from 0.78~(ASTE) to 0.86~(ACSA), reflecting task complexity. Cohen's Kappa indicates moderate-to-substantial agreement, peaking at $\kappa{=}0.65$ for Clause-Level, where A2 had adapted after completing ACSA first. ASTE yields the lowest agreement ($\kappa{=}0.52$, F1$\,{=}\,0.50$), reflecting the inherent difficulty of extracting complete triplets.

\textbf{Discussion.} Direct performance comparisons between ACAT and generic platforms are complicated by hidden data engineering costs (answering RQ5). Benchmarking raw annotation speed on a generic tool ignores the effort of writing custom ETL pipelines, reconstructing ASTE triplets from flat exports, and computing IAA externally. While ACAT achieves these results out of the box, equivalent figures on a generic platform would require post-hoc engineering.

\textbf{Limitations.} The evaluation uses two annotators on a single domain (restaurant reviews), limiting generalisation across annotator counts, text types, and domains. Fleiss' Kappa, implemented for ${\geq}3$ annotators, was not exercised. Scalability and concurrent-load benchmarks, as well as formal usability studies, remain as future work. The platform does not support all ABSA tasks. The interaction model is optimised for desktop.

\section{Conclusion} \label{sec:conclusions}
We introduced ACAT, an end-to-end data curation platform for ABSA. A preliminary experiment on $1{,}002$ restaurant reviews demonstrates efficient annotation (median 31.58~s) with moderate-to-substantial IAA. Revisiting our research questions:
\textbf{(RQ1)}~ACAT natively integrates four ABSA workflows without pre-annotation configuration.
\textbf{(RQ2)}~An automated ETL pipeline performs row-level alignment at export.
\textbf{(RQ3)}~The Implicit Toggle captures unstated aspects and sentiments.
\textbf{(RQ4)}~IAA metrics are computed natively during export.
\textbf{(RQ5)}~Direct comparisons with generic tools are unreliable because they ignore hidden ETL costs; ACAT handles the end-to-end pipeline natively.

\textbf{Future work} includes the official platform release, LLM-based pre-annotations to reduce time and cost, support for additional ABSA sub-tasks, cross-domain validation with more annotators, scalability benchmarks, and a real-time analytics dashboard.

\smallskip
\noindent\textbf{Acknowledgments: } The research presented in this paper was supported in part by (1) The Academy of Romanian Scientists, through the funding of the project ``NetGuardAI: Intelligent system for harmful content detection and immunization on social networks'' (AOȘR-TEAMS-IV); and
(2) the National University of Science and Technology, POLITEHNICA Bucharest, through the PubArt program.

\bibliographystyle{splncs04}
\bibliography{references}
\end{document}